\documentclass[letterpaper,10pt,conference]{ieeeconf}

\IEEEoverridecommandlockouts
\usepackage{hyperref}
\hypersetup{
    colorlinks = true,
    linkcolor = red,
    urlcolor= blue,
    pdftitle={DVS of TDCRs},
}
\usepackage{amsmath}

\usepackage{caption}
\usepackage{subcaption}
\usepackage{graphicx}
\usepackage{multirow}
\usepackage{cite}
\usepackage{color}
\usepackage{tabularx}
\usepackage{soul,color}

\title{\LARGE \bf Deep Direct Visual Servoing of Tendon-Driven Continuum Robots}

\author{Ibrahim Abdulhafiz$^{1}$, Ali A. Nazari$^{2,*}$, Taha Abbasi-Hashemi$^{1}$, Amir Jalali$^{2}$, \\Kourosh Zareinia$^{2}$, Sajad Saeedi$^{2}$, and Farrokh Janabi-Sharifi$^{2}$%
\thanks{$^{*}$Corresponding author; {Email: \href{mailto:ali.nazari@ryerson.ca}{ali.nazari@ryerson.ca}}}
\thanks{$^{1}$Department of Electrical, Computer, and Biomedical Engineering,
        Ryerson University, Toronto, ON M5B 2K3, Canada.}%
\thanks{$^{2}$Department of Mechanical and Industrial Engineering,
        Ryerson University, Toronto, ON M5B 2K3, Canada.}%
\thanks{This work was supported in part by the Natural Sciences and Engineering Research Council of Canada (NSERC) under Discovery Grants 2017-06930 and 2019-05562 and the Ryerson Faculty of Engineering and Architectural Science Dean's Research Fund (FEAS DRF).}%
}

\begin{document}

\maketitle

\begin{abstract}
Vision-based control provides a significant potential for the end-point positioning of continuum robots under physical sensing limitations. Traditional visual servoing requires feature extraction and tracking followed by full or partial pose estimation, limiting the controller's efficiency. We hypothesize that employing deep learning models and implementing direct visual servoing can effectively resolve the issue by eliminating such intermediate steps, enabling control of a continuum robot without requiring an exact system model. This paper presents the control of a single-section tendon-driven continuum robot using a modified VGG-16 deep learning network and an eye-in-hand direct visual servoing approach. The proposed algorithm is first developed in Blender software using only one input image of the target and then implemented on a real robot. The convergence and accuracy of the results in normal, shadowed, and occluded scenes demonstrate the effectiveness and robustness of the proposed controller.
\end{abstract}

\section{Introduction} \label{sec:intro}
Continuum robots (CRs) have become popular in recent years due to their continuum structure and compliance, enabling them to manipulate geometrically complex objects and work in unstructured and confined environments \cite{camarillo2008mechanics}. In particular, tendon-driven CRs have typically small diameter-to-length ratios \cite{amanov2019tendon}, presenting great potential for their navigation in confined spaces such as body cavities \cite{burgner2015continuum}. 
\begin{figure}
\begin{center}
    \begin{subfigure}[b]{.95\columnwidth}
        \includegraphics[width=\linewidth]{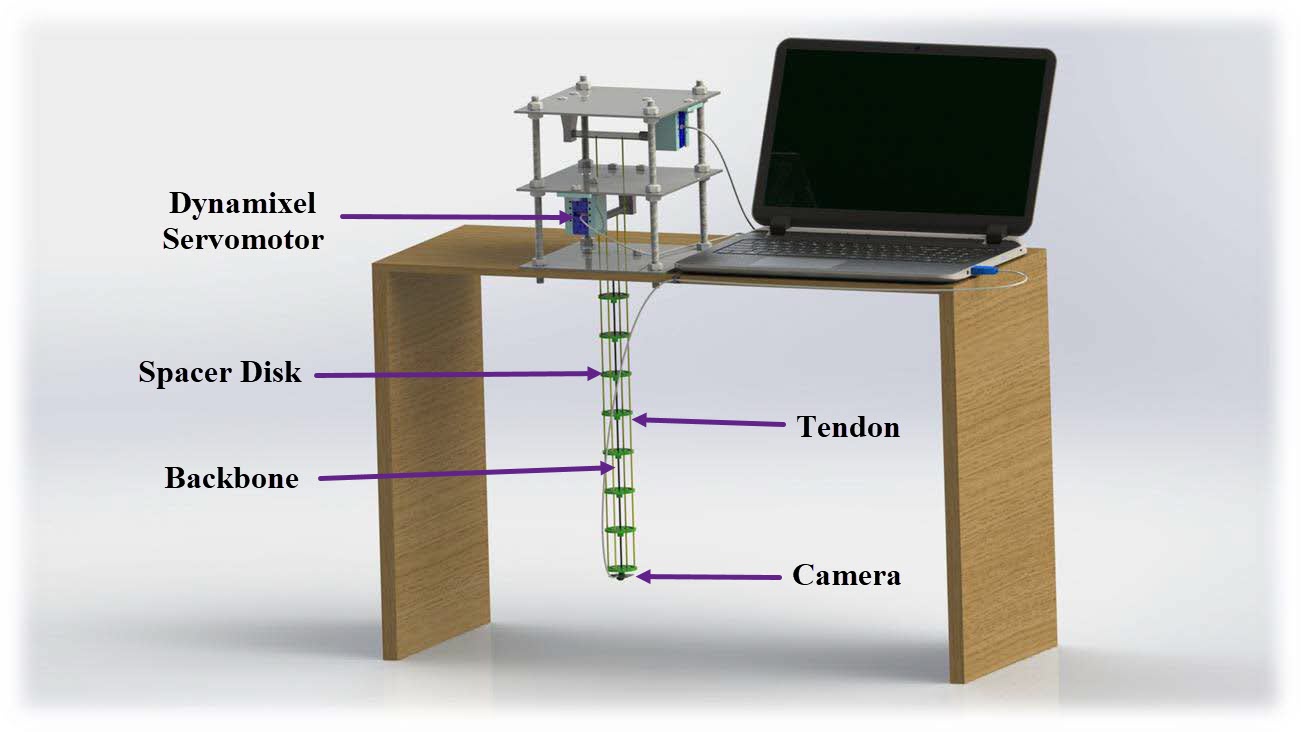}
        \caption{}
        \label{fig:CAD}
    \end{subfigure}
    \begin{subfigure}[b]{.9\columnwidth}
        \includegraphics[width=\linewidth,height=5cm]{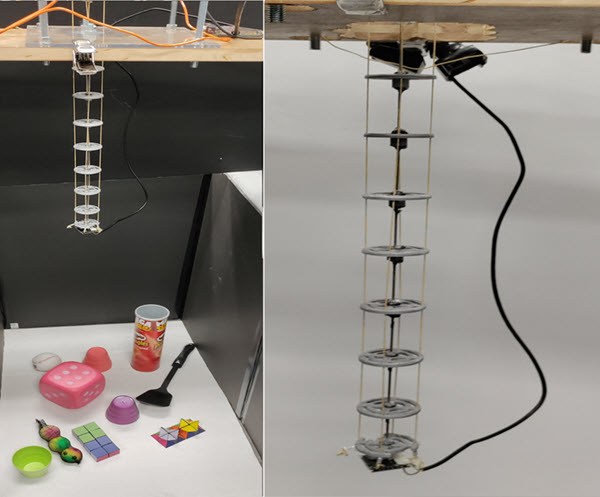}
        \caption{}
        \label{fig:TDCR}
    \end{subfigure}
    \caption{(a) CAD model and (b) prototype of tendon-driven continuum robot. A sample of robot motion can be seen in the supplemented video.}
\end{center}
\end{figure}

Nevertheless, affected by the intrinsic compliance and the high number of degrees of freedom (DOFs), the control of CRs has been a challenge since their emergence. Both model-based \cite{chikhaoui2018control} and model-free \cite {george2018control,da2020challenges} control approaches have been proposed. The issues related to modeling and sensing have contributed to the control challenge of CRs. Kinematic and dynamic modeling of CRs are ongoing research problems and often involve the iterative solutions of partial differential equations (PDEs) \cite{till2019real,janabi2021cosserat}, which are often contaminated with significant parameter uncertainties. Additionally, sensing presents its own set of challenges. For example, limitations due to the size, bio-compatibility, and sterilizability of common sensors limit their integration into CRs for many medical interventions \cite{nazari2021image}. Non-contact sensing methods such as vision-based techniques have thus found an important place in many interventions with CRs. In particular, imaging modalities are readily available in many medical interventions, circumventing the need for extra sensor integration \cite{fallah2020depth}. Therefore, vision-based control methods provide attractive solutions by enabling the use of feasible sensing, and also direct end-point control of CRs with potential robustness to modeling uncertainties related to the robot and target object \cite{fallah2020depth,nazari2022visual,hutchinson1996tutorial,janabi2010comparison}. 

Early methods of vision-based control, also called visual servoing (VS), relied on the image projection of geometric features such as points, lines, corners, edges, ridges, and blobs. Both eye-in-hand (EIH) and eye-to-hand (ETH) configurations were utilized \cite{hutchinson1996tutorial}. Depending on the nature of error used in the control law, two basic approaches in VS have been realized, which include image-based visual servoing (IBVS) and position-based visual servoing (PBVS) \cite{janabi2010comparison}. Examples of VS approaches to CR control include the work of Wang \textit{et al.} \cite{wang2013visual,wang2016visual} who selected the IBVS-EIH approach for kinematic control of a cable-driven soft robot. They developed an adaptive PD tracking controller by knowing the intrinsic and extrinsic camera parameters, but they did not consider visual sensing accuracy in the modeling. Zhang \textit{et al.} \cite{zhang2017visual} modeled the statics of a cable-driven parallel soft robot and implemented an open-loop/closed-loop switching controller in an IBVS-ETH scheme. The open-loop controller was developed based on finite element modeling in the simulation environment, which is computationally inefficient for real-time control. System and imaging constraints were formulated in an IBVS-based visual predictive control with ETH configuration developed for tendon-driven CRs \cite{fallah2020depth}. Model-less optimal feedback control in the IBVS-ETH scheme was used by Yip \textit{et al.} \cite{yip2014model} to control a planar tendon-driven CR in the task space. They estimated the image Jacobian using backward differencing.

Despite having several advantages, IBVS and PBVS methods have some disadvantages/limitations \cite{hutchinson1996tutorial,janabi2010comparison}. The real-time pose estimation in PBVS schemes is always a challenge. Another significant challenge for feature-based methods is the extraction of image features, which may require camera pose measurement, robust feature extraction, feature matching, and real-time tracking, all of which are complex and computationally expensive \cite{marchand2005feature}. The success of the feature-based visual servoing, in fact, depends on the tracking success and performance, \textit{i.e.}, the speed, accuracy, robustness, and redundancy of the visual features \cite{ourak2019direct}. 

Using non-geometric VS or direct visual servoing (DVS) is an alternative to eliminate the feature tracking requirement. For instance, Photometric VS \cite{collewet2011photometric} is a solution to the problem in a 2D scenario. It exploits the full image as a whole, uses the luminance of all pixels in the image, and avoids extracting geometric features of the image. Due to the redundancy of visual information in the control loop, DVS schemes are more accurate and robust than geometric feature-based VS methods \cite{duflot2019wavelet}. Although these methods do not require feature extraction, their convergence is inferior to that of the classical VS methods \cite{bateux2017visual}.

Deep learning methods have been recently proposed to tackle the issues mentioned above. Examples include the work of Bateux \textit{et al.} \cite{bateux2017visual,bateux2018training}, which is based on training a convolutional neural network (CNN) using images captured from different scenes of a target object along with their corresponding poses. The estimated poses were used in a resultant PBVS scheme to achieve real-time control of a rigid-link manipulator. The proposed method showed satisfactory results in both tested and unforeseen scenes \cite{bateux2018training}. Also, Felton \textit{et al.} \cite{felton2021training} proposed a deep network for end-to-end DVS in which the velocity of a camera mounted on a robot tip is predicted using a Siamese network. They trained the algorithm on a subset of the ImageNet dataset and tested its performance on a 6-DOF rigid-link robot. In spite of these studies, no study has been reported on investigating the DVS of CRs. There are significant challenges associated with CRs, which make their end-to-end DVS different and challenging. Examples include significant differences between kinematic and dynamic models from their rigid-link counterparts and the existence of considerable uncertainties associated with their models.

The objective of this paper is to develop the first deep learning-based end-to-end control of CRs utilizing DVS methods and its implementation in actuation space. Our contribution is as follows:
\begin{itemize}
    \item Developing a deep learning-based direct VS algorithm. The deep network is structured by modifying the VGG-16 network. The model is then trained using a self-provided dataset (generated by Blender software), which includes variations of only one target image with normal conditions, illumination changes, and occlusions.
    \item Conducting extensive simulation studies in Blender in normal and perturbed conditions and then evaluating the controller's performance on a real robot. The algorithm is experimentally validated in a variety of scenarios including the normal operation of the robot within the full range of its workspace. The robustness is also analyzed against variations in the lighting in the environment and partial occlusion. Finally, our approach is compared with a classical IBVS approach.
\end{itemize}

\section{Methodology} \label{sec:method}
There exist many challenges in implementing VS on CRs. Unlike rigid robots with stable designs and well-defined kinematic models, the flexibility and soft nature of CRs make them susceptible to various modeling inaccuracies and extremely sensitive to noise and disturbance. Examples of modeling uncertainties include extreme hysteresis, backlash, dead zone, and high sensitivity to disturbance. Therefore, regressing the desired camera velocity is not sufficient for accurate control of CRs. To address these uncertainties, we propose a joint space VS scheme to localize the end-effector at a target image frame. This is accomplished by implementing an end-to-end deep learning model that directly computes the desired tendon velocities from camera images. In order to train the model robustly, a simulation environment is created to generate an appropriate training dataset.

Our methodology is based on employing a deep learning network that has been already trained but repurposing it by changing the last layer and tailoring it for the desired task. Using the image frames captured in real time by a camera, the network produces the raw $\Delta q^*$ commands that can direct the robot to the desired target after a subsequent scaling by a proportional controller. The intended network is trained using a user-generated dataset of RGB images produced utilizing Blender software. The performance of the proposed method is evaluated through extensive simulation and experimental studies in normal and changing conditions to prove that the algorithm is robust against lighting changes and partially occluded environments. 

\subsection{Prototype Design and Development}
As shown in Fig. \ref{fig:CAD}, the prototype CR has one section comprised of a flexible backbone made of spring steel, four braided Kevlar lines (Emmakites, Hong Kong) with a diameter of 0.45 mm as tendons, and spacer disks to route the tendons. The tendons were placed around the backbone with an offset of 1.8 mm, and an angular distance of $90^{\circ}$ from each other. The tendons were routed toward the robot tip by eight equally distanced spacer disks, which were 3D printed using PLA filament. The disks were solidly attached to the backbone using steel-reinforced epoxy adhesive with a strength of 3960 psi. A custom fixture was 3D printed to rigidly mount a 1080P HD webcam (OURLINK, CA, USA) on the robot tip in an EIH mode. The fixture was screwed on the last spacer disk such that it guarantees the minimum space between the camera and the robot tip while having no contact between the fixture and the camera's electronic board. The tendons were actuated using Dynamixel AX-12A servomotors (Robotis, CA, USA). Table \ref{tab:CR_design} provides the prototype specifications.
\begin{table}
\caption{Prototype specifications.}
\label{tab:CR_design}
\begin{center}
    \begin{tabularx}{\columnwidth}{l l l}
    \hline
    \textbf{Prototype's Part} & \textbf{Specification} & \textbf{Value} \\
    \hline
    \multirow{4}{*}{Backbone} & Density ($\rho$) & 7800\, Kg/m$^3$ \\
     & Young's modulus ($E$) & 207\, GPa \\
     & Length ($L$) & 0.4 m \\
     & Radius ($r$) & 0.9 mm \\
    \hline
    Tendon & Breaking strength & 31.75 Kg \\
    \hline
    \multirow{3}{*}{USB webcam} & Frame rate & 30\, fps \\
     & Resolution & 1080$\times$7200 pixels \\
     & Field of view (FOV) & 19$^{\circ}$ \\
     \hline
    \end{tabularx}
\end{center}
\end{table}

\subsection{Control Law}
\begin{figure}
\begin{center}
    \includegraphics[width=.95\columnwidth]{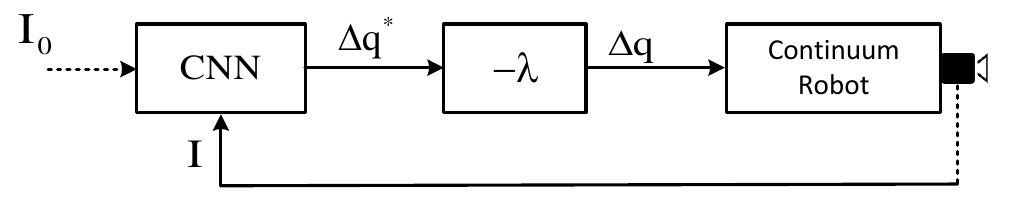}
    \caption{Block diagram of the proposed visual servo controller comprised of a camera in an EIH mode, a tendon-driven CR, and a CNN model.}
    \label{fig:ctrlDiagram}
\end{center}
\end{figure}

Classical VS approaches require complex Jacobian mapping, which is difficult to derive. Therefore, we aim to replace the entire mapping from image space to joint space with a learned model, such that the error between the current image frame, $I$, and the desired image frame, $I_0$, be minimized. As shown in Fig. \ref{fig:ctrlDiagram}, the output is multiplied by a gain, $-\lambda$, and fed into the CR. The control law is then stated as
\begin{align}
    \Delta q^* = f(I_0,I) \\
    \Delta q = -\lambda\,\, {\Delta q^*}
    \label{eq:ctrlLaw}
\end{align}
where $\Delta q$ and $\Delta q^*$ are respectively the change and the desired change in tendon displacement in mm, $I_0$ is the target image, $I$ is the current image, and $f()$ is a function of the target and the current images implemented on a modified VGG-16 network to output $\Delta q^*$.

\subsection{Neural Network Design}
\begin{figure}
\begin{center}
    \includegraphics[width=\columnwidth]{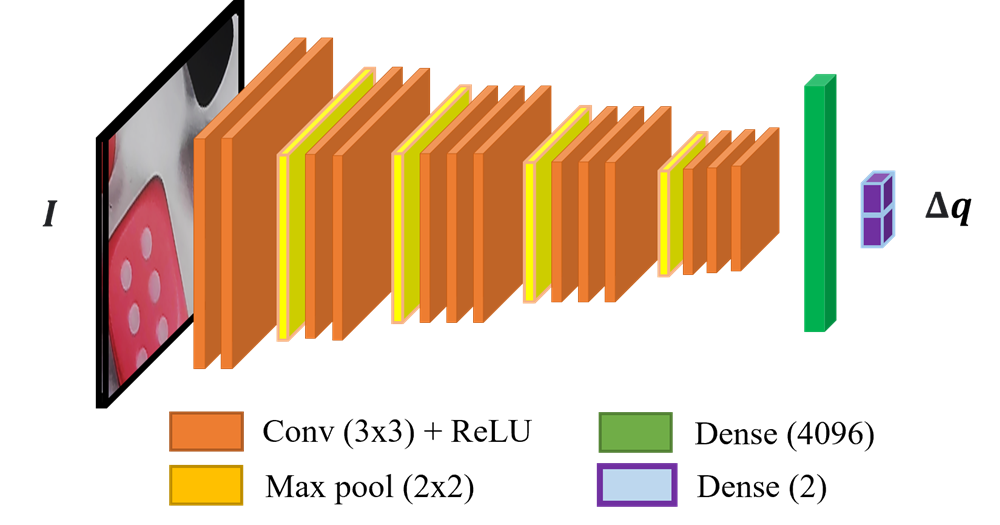}
    \caption{Architecture of modified VGG-16 model.}
    \label{fig:Model}
\end{center}
\end{figure}

In order to create an efficient neural network for our purpose, we utilized a VGG-16 backbone pre-trained on ImageNet to facilitate transfer learning \cite{vgg}. Having been trained on natural images, only the lower layers of the network will need to be trained to regress desired tendon displacements. In our model, the first 10 layers were frozen to speed training because they already contained low-level features from natural images. We modified this network by dropping out the last dense layer and replacing it with a dense layer with two outputs corresponding to $q_1$ and $q_2$. The activation function was set to linear (see Fig. \ref{fig:Model}). Various alternatives were considered to challenge this proposed model. Firstly, different backbones were considered, particularly ResNet50 and ResNet101. Secondly, we considered adding multiple dense layers to improve the nonlinear fitting of the CR model.

\subsection{Simulation Environment}
Training in the simulation provides various advantages over the real world. Not only are they much quicker in acquiring the data, but they also offer the opportunity to structure the environment to account for various noises and uncertainties, enabling the model to be more robust. On the contrary, attempting to learn the dynamics and uncertainties of the robot remains challenging in simulated environments. We resolved this problem by utilizing an open-source 3D computer graphics software called Blender and creating an environment that models the pose of the end-effector given a tendon displacement, $q$, value. This was achieved by using the forward kinematics of the robot to place and orient the virtual camera in the simulated environment. Being a single-section 2-DOF CR, we modeled the kinematics based on the constant curvature assumption, as presented by Rao \textit{et al.} \cite{rao2021model}.

\begin{figure}
\begin{center}
    \begin{subfigure}[b]{0.24\columnwidth}
        \includegraphics[width=\linewidth]{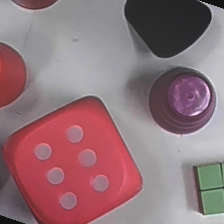}
        \caption{}
    \end{subfigure}
    \begin{subfigure}[b]{0.24\columnwidth}
        \includegraphics[width=\linewidth]{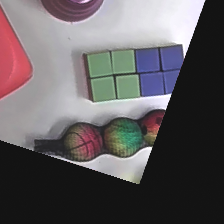}
        \caption{}
    \end{subfigure}
    \begin{subfigure}[b]{0.24\columnwidth}
        \includegraphics[width=\linewidth]{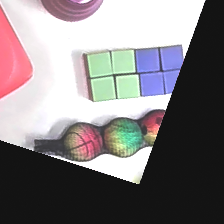}
        \caption{}
    \end{subfigure}
    \begin{subfigure}[b]{0.24\columnwidth}
        \includegraphics[width=\linewidth]{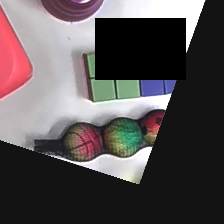}
        \caption{}
    \end{subfigure}
    \caption{Typical set of images used in simulation. (a) target image, (b) camera view at the tendon displacement of $(q_1,q_2) = (4,-3)\, mm$, (c) camera view with random lighting, and (d) camera view with occlusion.}
    \label{fig:sampleImages}
\end{center}
\end{figure}

Whereas this approach ignores the dynamic effects of the CR, we propose that implementing robust vision control would allow the feedback loop to correct for most of the aforementioned challenges of the CR. Shadowing and occlusion were included to provide this robustness. Shadowing was achieved by adjusting the light source in the environment, whereas occlusion was achieved by placing black rectangles of random positions and dimensions within the image. Fig. \ref{fig:sampleImages} shows some samples from the simulation.

\begin{figure}
\begin{center}
    \includegraphics[width=.95\columnwidth,height=4cm]{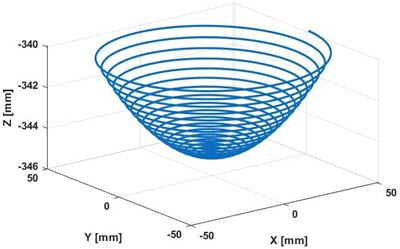}
    \caption{Camera path used to generate the dataset. Assuming the camera is exactly mounted on the robot's tip, $(x,y,z)$ are the coordinates of the camera's center point with respect to the robot's base.} 
    \label{fig:spiral}
\end{center}
\end{figure}

For acquiring the dataset, previous approaches made use of two sets of images; one randomly placed within a distance from the origin for general convergence, and the other one very close to the origin for fine-tuning \cite{bateux2018training}. We thought this binary approach would produce noisier joint commands, and therefore we implemented a continuous method. The farther away the CR is from the origin, the sparser the dataset will be. Similarly, to produce more deterministic results, a spiral path was used to traverse all reaching points in the 3D world within a certain threshold. The intended path can stimulate nonlinearities of the robot very well while covering all quadrants of the robot's workspace even if there is no overlap, which is of our interest in experiments. The spiral path was generated using
\begin{align}
    {q_1} = \frac{A}{n}x\,\cos\, (2\pi \frac{P}{n}x\,)
    \label{eq:q1}
\end{align}
\begin{align}
    {q_2} = \frac{A}{n}x\,\sin\, (2\pi \frac{P}{n}x\,)
    \label{eq:q2}
\end{align}
where $A$ is the maximum displacement of the tendon, $P$ is the total number of periods the CR makes, $n$ is the number of sample points, and $x$ is an integer from $1$ to $n$. Fig. \ref{fig:spiral} shows the generated spiral path.
Using Blender, we created the environment and overlaid the desired scene. Employing a Python API, we moved the camera and light source around and captured images from the scene for training purposes.

The benefits of using a simulation environment instead of generating the dataset from the physical robot are twofold. Firstly, only one target image is required, and thus neither an exact camera is needed nor is having access to the physical robot required. So long as simple forward kinematics is known, Blender can be used to generate the entire dataset. This may be true, especially in medical procedures whereby only limited pre-operative image data is present in advance. Secondly, the use of simulation enables the dataset to be made robust to various lighting, noises, artifacts, and other disturbances.

\subsection{Training and Validation}
For training, we selected mean squared error (MSE) as the loss function because we designed our output activation function as linear. This was because we wanted to learn a one-to-one mapping of the ground truth control points. The ground truth of the dataset was generated based on equation (\ref{eq:tanh}), which keeps the ground truth between $-1$ and $1$, allowing for better training. Similarly, this mapping produces smoother convergence profiles (as opposed to linear mapping).
\begin{align}
    q_{mapped} = \tanh \left(10\,q \right)
    \label{eq:tanh}
\end{align}

In comparison, we linearly mapped $-5\,mm$ and $5\,mm$ to $-1$ and $1$, respectively, clipping any values beyond. However, such an approach would not penalize the optimizer as much near the origin. Since we aim to propose sub-millimeter accuracy, utilizing equation (\ref{eq:tanh}) would force convergence while producing a much smoother velocity profile.

The simulation environment was used to generate the dataset. To this end, 5000 images were acquired with a maximum amplitude of $7\,mm$ and a period of 20. Random lighting effect and random occlusion were included. These occlusions were represented as black rectangles overlaid at random positions to force the model to learn the full spatial features and make it more robust. As we used the classical VGG-16, the input image was RGB of size $224\times224$. The model was trained for 50 epochs with a batch size of 32 and a learning rate of $1e-5$ using Adam optimizer. The final MSE was determined to be $3e-5$. To validate our hypothesis, VGG-16 was swapped with ResNet51 and ResNet101. As expected, the training took substantially longer, and the MSE was inferior to that of the VGG. Similarly, two dense layers (1024 and 512, respectively) were added between the VGG and the final $q$ output to test our hypothesis. Nonetheless, the training took longer without any significant improvement to the MSE. Training on an Nvidia Titan Xp GPU was reasonably fast, taking less than 20 minutes on 50 epochs to train. The model inference was also extremely fast, taking about $15\, ms$ per frame\footnote{The code and dataset will be made available publicly once the paper is accepted.}.

\section{Experimental Results} \label{sec:exp}
The effectiveness and efficiency of the developed controller were tested in a variety of simulations and experiments. Here, we describe the simulations conducted using Blender software in order to test the robustness and accuracy of the controller. Following this, experimental studies are discussed in more detail, including the experiment design to cover all four quadrants of the robot workspace, as well as a discussion of the accuracy of the results. Finally, the deep learning-based controller is compared to a classical IBVS controller to verify that the obtained results from CNN-based VS are significant compared to the classical one.

\subsection{Simulation}
Before conducting experiments in real-world scenarios, we performed some tests to validate the robustness and accuracy of the simulation. Since the kinematics did not incorporate nonlinear effects when generating the dataset, we needed to include various uncertainties to prove the model's robustness within the simulation.

\subsubsection{Modeling Uncertainties}
Since the constant curvature assumption does not hold true in all situations, other uncertainties and disturbances were added to the simulation. Regarding geometric uncertainties, derived from parameters of the robot including length, disk space, etc., Gaussian noise with a mean of $0$ and standard deviation of $0.01\, mm$ was added to the output $q$ values. Also, the outputs of the trained model were scaled to uniformly distributed random numbers in the range of $0.25$ to $4$. Random lighting was introduced to account for the vision uncertainties, and a region within the image was occluded with black rectangles. Instead of generating these random scene environments every iteration, we chose to regenerate these random uncertainties every 20 iterations to better model the changing conditions of the real-world environment.

\subsubsection{Simulation Results}
\begin{figure}
\begin{center}
    \begin{subfigure}[b]{0.24\columnwidth}
        \includegraphics[width=\linewidth]{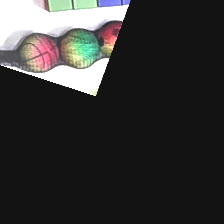}
        \caption{}
    \end{subfigure}
    \begin{subfigure}[b]{0.24\columnwidth}
        \includegraphics[width=\linewidth]{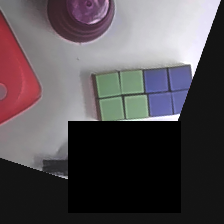}
        \caption{}
    \end{subfigure}
    \begin{subfigure}[b]{0.24\columnwidth}
        \includegraphics[width=\linewidth]{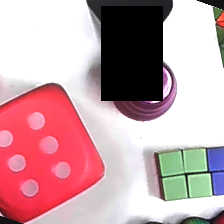}
        \caption{}
    \end{subfigure}
    \begin{subfigure}[b]{0.24\columnwidth}
        \includegraphics[width=\linewidth]{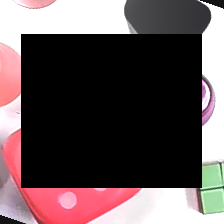}
        \caption{}
    \end{subfigure}
    \begin{subfigure}[b]{0.24\columnwidth}
        \includegraphics[width=\linewidth]{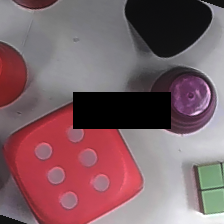}
        \caption{}
    \end{subfigure}
    \begin{subfigure}[b]{0.24\columnwidth}
        \includegraphics[width=\linewidth]{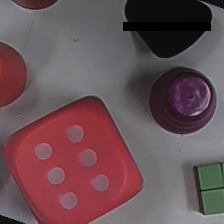}
        \caption{}
    \end{subfigure}
    \begin{subfigure}[b]{0.24\columnwidth}
        \includegraphics[width=\linewidth]{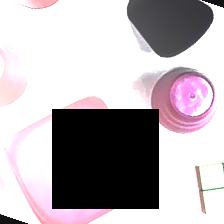}
        \caption{}
    \end{subfigure}
    \begin{subfigure}[b]{0.24\columnwidth}
        \includegraphics[width=\linewidth]{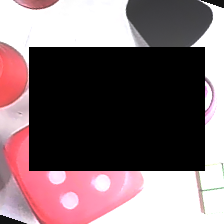}
        \caption{}
    \end{subfigure}
    \caption{Sequence of camera views in a typical simulation started at the tendon displacement of $(q_1,q_2) = (6,-4)\, mm$ at iteration numbers of (a) 1, (b) 25, (c) 50, (d) 75, (e) 100, (f) 225, (g) 250, and (h) 299.}
    \label{fig:sim_sequence}
\end{center}
\end{figure}

\begin{figure}
\begin{center}
    \setlength{\fboxsep}{0pt}%
    \setlength{\fboxrule}{0.1pt}%
    \fbox{\includegraphics[width=.95\columnwidth]{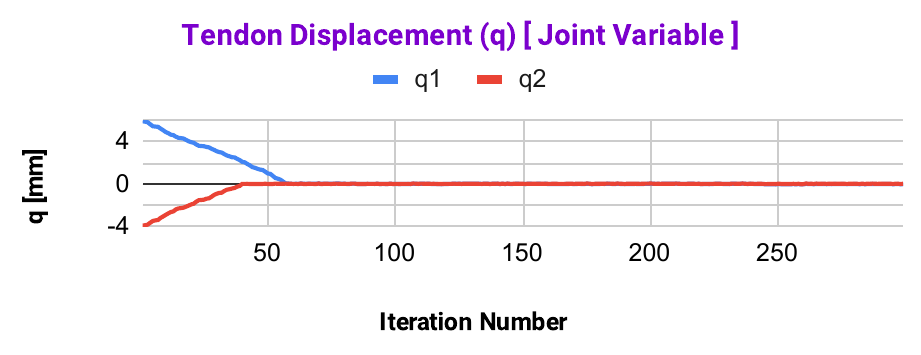}}
    \fbox{\includegraphics[width=.95\columnwidth]{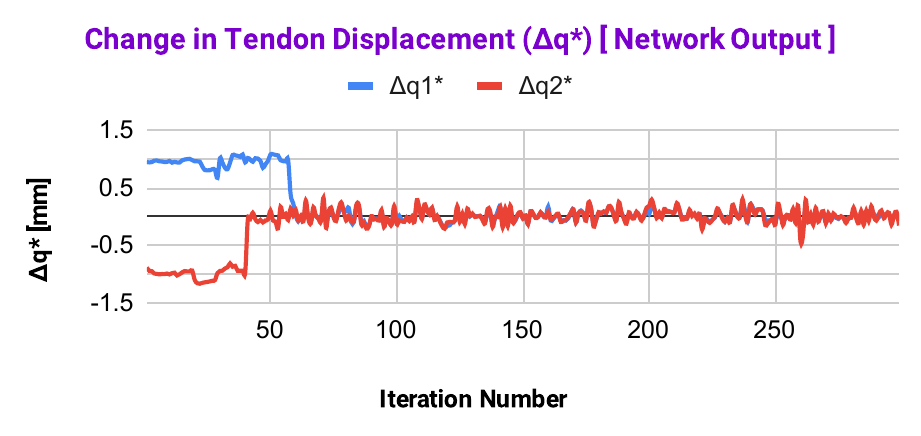}}
    \caption{Tendon displacement (top) and raw $\Delta q^*$ commands from the network, before being multiplied by $-\lambda$,  to stabilize the robot (bottom).}
    \label{fig:out}
\end{center}
\begin{center}
    \begin{subfigure}[b]{0.24\columnwidth}
        \includegraphics[width=\linewidth]{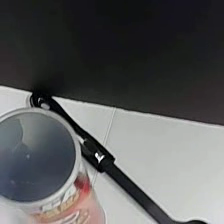}
        \caption{}
    \end{subfigure}
    \begin{subfigure}[b]{0.24\columnwidth}
        \includegraphics[width=\linewidth]{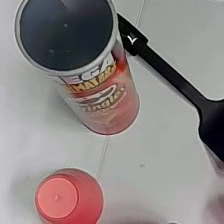}
        \caption{}
    \end{subfigure}
    \begin{subfigure}[b]{0.24\columnwidth}
        \includegraphics[width=\linewidth]{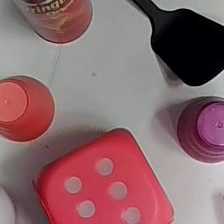}
        \caption{}
    \end{subfigure}
    \begin{subfigure}[b]{0.24\columnwidth}
        \includegraphics[width=\linewidth]{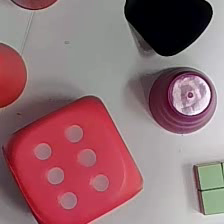}
        \caption{}
    \end{subfigure}
    \begin{subfigure}[b]{0.24\columnwidth}
        \includegraphics[width=\linewidth]{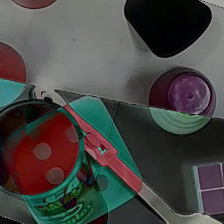}
        \caption{}
    \end{subfigure}
    \begin{subfigure}[b]{0.24\columnwidth}
        \includegraphics[width=\linewidth]{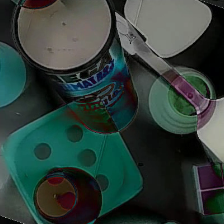}
        \caption{}
    \end{subfigure}
    \begin{subfigure}[b]{0.24\columnwidth}
        \includegraphics[width=\linewidth]{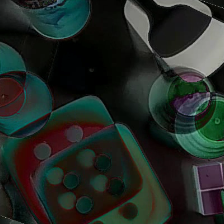}
        \caption{}
    \end{subfigure}
    \begin{subfigure}[b]{0.24\columnwidth}
        \includegraphics[width=\linewidth]{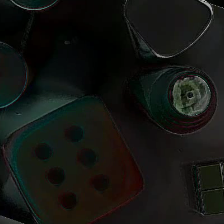}
        \caption{}
    \end{subfigure}
    \caption{(a-d) Sequence of camera views in a typical experiment started at the tendon displacement of $(q_1,q_2) = (5,-7)\, mm$, (e-h) Corresponding difference between the normalized target and intended images.}
    \label{fig:exp_main_imgs}
\end{center}
\end{figure}

Simulating with the initial tendon displacements of $(q_1,q_2) = (6,-4)\, mm$ we noticed the CR is able to converge smoothly although less than $25\%$ of the target image was visible at the starting position. Moreover, the change in lighting, as shown in Fig. \ref{fig:sim_sequence}, did not affect the convergence of the CR. More interestingly, adding occlusion (at times greater than $80\%$) did not destabilize the CR and, as noted with the raw network output in Fig. \ref{fig:out}, was still able to counteract the Gaussian noise added to the actuation commands of the CR. A sample of simulation studies can be seen in the supplemented video. Having been successfully validated in simulation, the next section will extend it to the real-world environment.

\subsection{Experimental Validation}
In order to test the practicality of the proposed end-to-end model, we applied it to the experimental setup developed for the purpose to show its robustness to various noises and uncertainties. Fig. \ref{fig:TDCR} shows the prototyped robot for the experiment.

\subsubsection{Experiment Design}
The physical environment was structured in a similar way to the simulation environment, as shown in Fig. \ref{fig:TDCR}. To test the model's accuracy in converging the CR, the end-effector was moved to random positions, and the trained model attempted to minimize the difference between the current image frame $I$ and the target image $I_0$ on which the model has been trained. The range of motion was limited to $\pm 10\, mm$ for each tendon to keep the scene within the camera's field of view (FOV). Note that the scene with which the model was trained was larger than the camera's FOV, which enabled our model to operate despite there being no overlap with the target image. Further, to demonstrate the robustness of the controller, the robot was operated under dynamic lighting conditions, dynamic occlusion, and finally partial static occlusion.

\subsubsection{Results and Discussion}
\begin{figure}
\begin{center}
    \setlength{\fboxsep}{0pt}%
    \setlength{\fboxrule}{0.1pt}%
    \fbox{\includegraphics[width=.95\columnwidth]{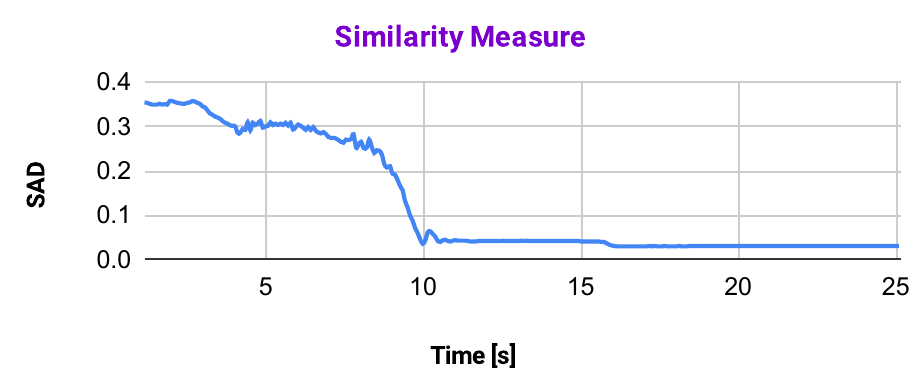}}
    \fbox{\includegraphics[width=.95\columnwidth]{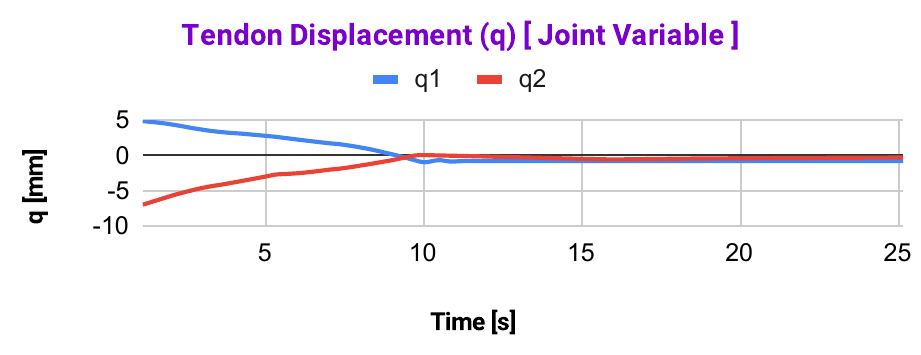}}
    \fbox{\includegraphics[width=.95\columnwidth]{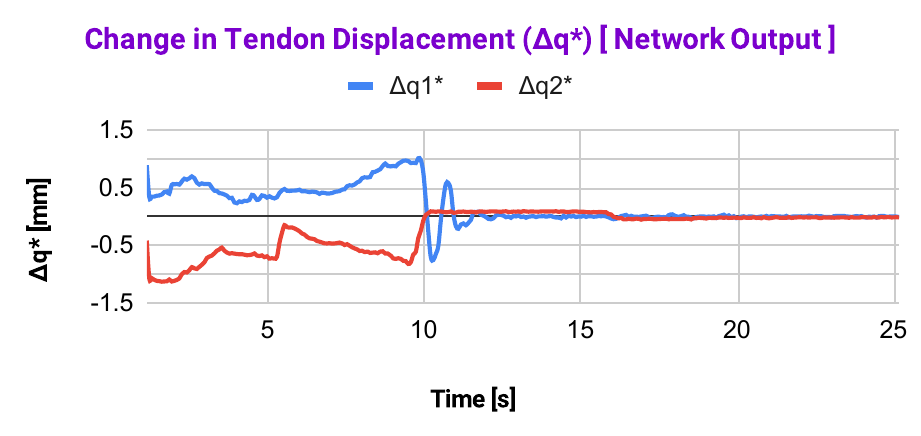}}
    \caption{SAD between $I^*$ and $I_0^*$ (top), tendon displacement at each iteration (middle), and the raw $\Delta q^*$ commands (bottom).}
    \label{fig:exp_main_plots}
\end{center}
\end{figure}

Experimenting from the initial tendon displacements of $(q_1,q_2) = (5,-7)\, mm$, the first row in Fig. \ref{fig:exp_main_imgs} shows that the CR converges to match the camera image to the target image. Normalizing and then subtracting the current and target images gives the images on the second row. Upon convergence, the overlap between the two images becomes highly precise. To evaluate the convergence quantitatively, the pixel-wise sum of absolute distance (SAD) between the normalized target and current images was calculated using
\begin{align}
    SAD = \sum |I^* - I_0^*|
    \label{eq:SAD}
\end{align}
where $I^*$ is the normalized current image and $I_0^*$ is the normalized target image. As shown in Fig. \ref{fig:exp_main_plots}, the SAD value fails to approach zero, stating that the lighting environment and image exposure were slightly different.
\begin{figure}
\begin{center}
    \begin{subfigure}[b]{0.3\columnwidth}
        \includegraphics[width=\linewidth,height=2cm]{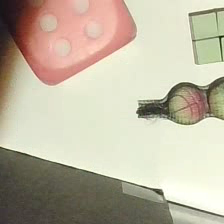}
        \caption{}
    \end{subfigure}
    \begin{subfigure}[b]{0.3\columnwidth}
        \includegraphics[width=\linewidth,height=2cm]{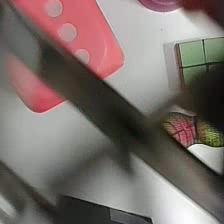}
        \caption{}
    \end{subfigure}
    \begin{subfigure}[b]{0.3\columnwidth}
        \includegraphics[width=\linewidth,height=2cm]{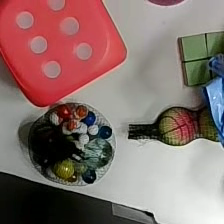}
        \caption{}
    \end{subfigure}
    \begin{subfigure}[b]{0.3\columnwidth}
        \includegraphics[width=\linewidth,height=2cm]{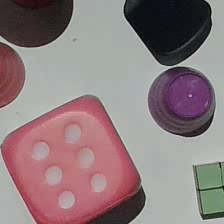}
        \caption{}
    \end{subfigure}
    \begin{subfigure}[b]{0.3\columnwidth}
        \includegraphics[width=\linewidth,height=2cm]{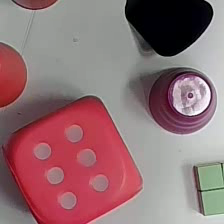}
        \caption{}
    \end{subfigure}
    \begin{subfigure}[b]{0.3\columnwidth}
        \includegraphics[width=\linewidth,height=2cm]{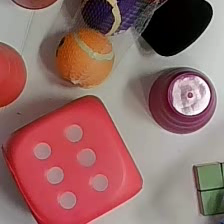}
        \caption{}
    \end{subfigure}
    \begin{subfigure}[b]{0.3\columnwidth}
        \includegraphics[width=\linewidth,height=2cm]{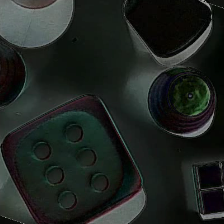}
        \caption{}
    \end{subfigure}
    \begin{subfigure}[b]{0.3\columnwidth}
        \includegraphics[width=\linewidth,height=2cm]{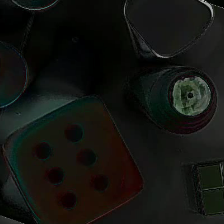}
        \caption{}
    \end{subfigure}
    \begin{subfigure}[b]{0.3\columnwidth}
        \includegraphics[width=\linewidth,height=2cm]{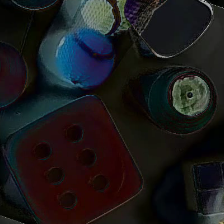}
        \caption{}
    \end{subfigure}
    \caption{(a-c) Initial views of the camera in the robustness analyses using dynamic lighting, dynamic occlusion, and partial static occlusion, respectively. (d-f) Corresponding converged views. (g-i) Corresponding difference between the target image and converged views.}
    \label{fig:robust_exp_img}
\end{center}
\end{figure}

Note that $q_1$ and $q_2$ values do not approach 0 in Fig. \ref{fig:exp_main_plots} despite that being the origin with which the target image was taken. This can be assigned to the dynamic effects of the CR and, more specifically, its hysteresis, which will be addressed in our future research. Without a computationally complex controller, our model can converge the camera frame to the target image by automatically compensating for the nonlinear effects. The raw $\Delta q$ commands show that the model smoothly converges to the origin and stabilizes once it approaches.

In order to test the robustness, different lighting conditions and occlusions were considered. As seen in Fig. \ref{fig:robust_exp_img}, the top row images show the CR at the starting position, $(q_1,q_2) = (-2,2)\, mm$. The left, center, and right columns show the dynamic lighting, dynamic occlusion, and static partial occlusion scenarios, respectively. The bottom row shows the corresponding normalized differences with the target image at the final iteration for each scenario. Despite the uncertainties, the precise overlap confirms that the model has learned sufficiently through simulation alone and that it can robustly control the CR in an end-to-end fashion. Note that even though the simulation only used one 2D image of the target scene, which results in projection artifacts when simulating in 3D, the model successfully operated in the real 3D environment without additional modifications. Four examples of experimental validation and robustness analysis can be seen in the supplemented video.

\subsection{Comparison with Classical VS}
To determine the superiority of our approach, we conducted similar experiments using classical VS approaches. One initial downside of classical VS, as opposed to our approach, is the requirement of all features to be visible in the camera frame at all times. Hence, we were restricted to tendon displacements of $(q_1,q_2) = (\pm 2, \pm 2)\, mm$ to make sure the features always remain in the camera's FOV. For testing purposes, template matching of the pink dice circles was used in the classical approach.

\subsubsection{Classical Control Law}
Classical IBVS aims to minimize the pixel error between the current and the target features. In our case, four feature points were selected using a template matching algorithm. Thereafter, binary thresholding was used to extract regions of high feature similarity. Finally, the centroids of these features were extracted, resulting in a $(u,v)$ coordinate for each of the features. Given four feature points, the classical image Jacobian for each feature was
\begin{align}
    L_x = \begin{bmatrix} 
	\frac{f}{z} & 0 & -\frac{u}{z} & -\frac{uv}{f} & \frac{f^2+u^2}{f} &  -v \\
	0 & \frac{f}{z} & -\frac{v}{z} & \frac{-f^2+v^2}{f} & \frac{uv}{f} & u\\
	\end{bmatrix}
\end{align}
where $f$ and $z$ are the camera's focal length and the image depth, respectively. The depth was considered constant, equal to 1 m. After calculating four Jacobian matrices corresponding to four features, the image Jacobian was found as
\begin{align}
    J_{img} = \begin{bmatrix} 
	L_{x1} \\
	L_{x2} \\
	L_{x3} \\
	L_{x4} \\
	\end{bmatrix}
    \label{eq:Jimg}
\end{align}
In order to enhance the computational efficiency, the Jacobian matrix of the robot was approximated using a finite difference method \cite{leibrandt2015line}. To this end, the change in the joint space variable, $\Delta q$, was set to be 0.1 mm, which is small enough for a submillimeter accuracy. Then, the final interaction matrix, $L_e$, was computed as 
\begin{align}
    L_e = J_{img} H J_{robot}
    \label{eq:L}
\end{align}
where 
\begin{align}
    H = \begin{bmatrix} 
	R_{3\times3} & 0_{3\times3} \\
	0_{3\times3} & R_{3\times3}\\
	\end{bmatrix}
    \label{eq:R}
\end{align}
and $R$ is the rotation matrix from the base frame to the camera frame. The final classical IBVS control law was
\begin{align}
    \dot{q} = -\lambda \, L_{e}^{+} (s - s^*),
    \label{eq:cvscontol}
\end{align}
where $\dot{q}$ is the velocity of the tendon, $\lambda$ is a gain factor, $L_{e}^{+}$ is the pseudo-inverse of the interaction matrix, $s$ is the current feature and $s^*$ is the target feature.

\subsubsection{Results and Discussion}
\begin{table}
\caption{Comparison between classical and CNN-based VS.}
\label{tab:classical_vs}
\begin{center}
    \begin{tabularx}{\columnwidth}{c l c c}
    \hline
    \textbf{Quadrant} & \textbf{Metrics} & \textbf{Classical} & \textbf{CNN-based} \\
    \hline
    \multirow{4}{*}{\#1 (+, +)} & Initial $q$ & $(2,0.5)$ & $(4,4)$ \\
     & Final SAD & \textbf{0.046} & 0.057 \\
     & Convergence (\# of iterations) & \textbf{101} & 191 \\
    \hline
    \multirow{4}{*}{\#2 (+, -)} & Initial $q$ & $(1,-1.5)$ & $(3,-5)$ \\
     & Final SAD & 0.065 & \textbf{0.058} \\
     & Convergence (\# of iterations) & 88 & \textbf{85} \\
    \hline
    \multirow{4}{*}{\#3 (-, -)} & Initial $q$ & $(-2,-2)$ & $(-2,-3)$ \\
     & Final SAD & \textbf{0.036} & 0.054 \\
     & Convergence (\# of iterations) & 105 & \textbf{60} \\
    \hline
    \multirow{4}{*}{\#4 (-, +)} & Initial $q$ & $(-1,2)$ & $(-3,6)$ \\
     & Final SAD & \textbf{0.027} & 0.060 \\
     & Convergence (\# of iterations) & 228 & \textbf{127} \\
    \hline
    \end{tabularx}
\end{center}
\end{table}

\begin{figure}
\begin{center}
    \setlength{\fboxsep}{0pt}%
    \setlength{\fboxrule}{0.1pt}%
    \fbox{\includegraphics[width=.95\columnwidth]{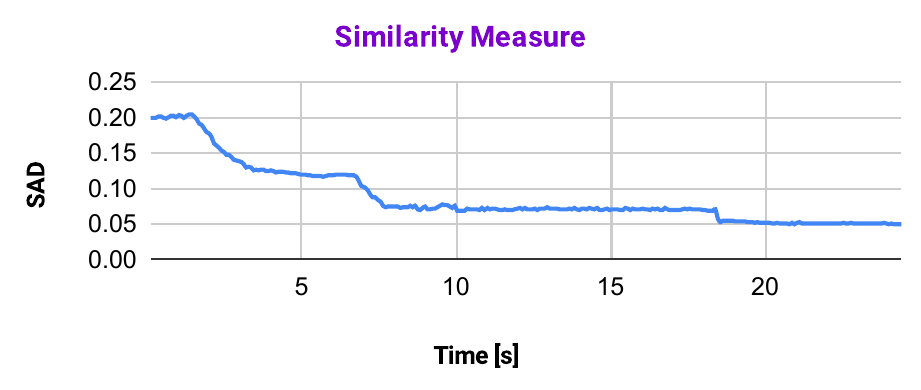}}
    \fbox{\includegraphics[width=.95\columnwidth]{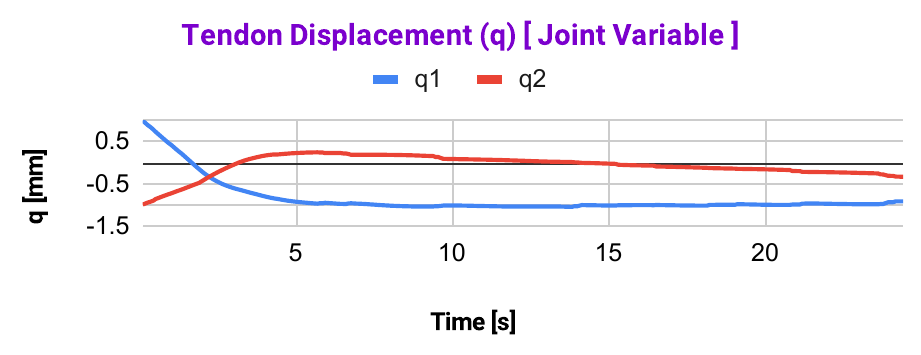}}
    \fbox{\includegraphics[width=.95\columnwidth]{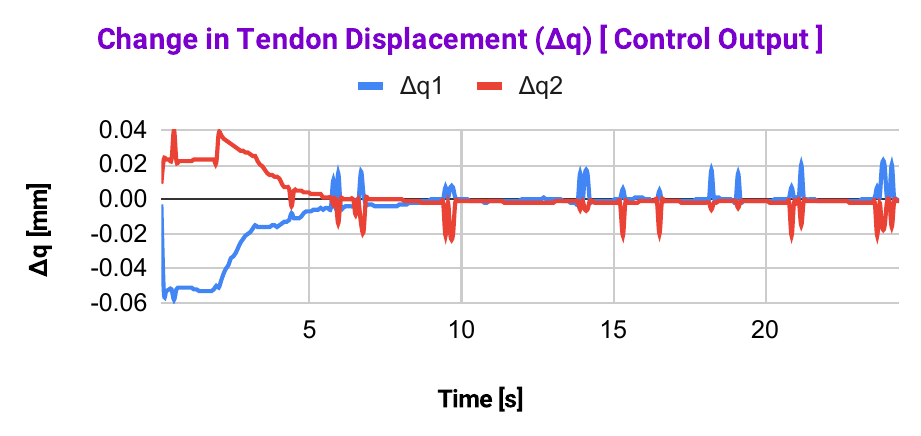}}
    \caption{Classical VS: SAD between $I^*$ and $I_0^*$ (top), tendon displacement at each iteration (middle), and the $\Delta q$ commands (bottom).}
    \label{fig:exp_classical_plots}
\end{center}
\end{figure}

As seen in Table \ref{tab:classical_vs}, our approach is more advantageous than the classical VS approach given its larger operating workspace. Despite having greater initial $q$ values, our approach is capable of producing similar and, at times, more superior results. We chose to start at different initial configurations within each quadrant to show the robustness of convergence for both approaches.

A primary advantage of our approach is that it takes into account the entire image rather than particular features, making it more robust to noise, partial occlusion, and artifacts. Our model can converge in situations where the initial and target images do not overlap, thereby expanding the servoing workspace. Secondly, our end-to-end approach does not require a Jacobian of the robot and hence produces smooth trajectories throughout the servoing. This is contrary to the classical VS which requires the robot Jacobian to be known. However, CRs are known to have extensive dynamic effects, making accurate computation of the Jacobian challenging. In our case, the robot Jacobian was derived using simple kinematics and therefore did not take into account various uncertainties present in the CR. These typically result in slower convergence time and a noisier displacement profile when compared to our deep learning-based approach, as evident from Fig. \ref{fig:exp_classical_plots}. Consequently, the gain needs to be set much smaller to produce a smoother trajectory profile, thereby resulting in the slower convergence time observed.

\section{Conclusion and Future Work} \label{sec:concl}
In this paper, a deep direct visual servoing algorithm was proposed to control a single-section tendon-driven continuum robot in an eye-in-hand configuration. The advantage of our training approach is using single image and populating the images utilizing Blender software for generating a dataset for VGG-16 network. The dataset includes different views of the scene plus illumination change and occlusion for robustness analysis. The algorithm was tested on a real robot developed by the team and showed fast and accurate convergence in regular scenes. Also, the algorithm's robustness was verified in scenarios incorporating dynamic illumination changes as well as dynamic and static occlusions.

In our future work, the robot will be extended to multiple sections, and the dynamic effects of the robot motion such as hysteresis, tendon slack, backlash, dead zone, and external disturbances will be investigated. We will also extend the control approach to achieve faster convergence and improved robustness to the aforementioned dynamic effects.

\bibliographystyle{IEEEtran}
\bibliography{Manuscript.bib}

\end{document}